\documentclass[10pt,twocolumn,letterpaper]{article}

\usepackage{wacv}
\usepackage{times}
\usepackage{epsfig}
\usepackage{graphicx}
\usepackage{amsmath}
\usepackage{amssymb}
\usepackage{mathrsfs}
\usepackage{array}
\usepackage{booktabs}
\usepackage{adjustbox}
\usepackage{caption}
\usepackage{cuted}
\usepackage{capt-of}
\usepackage[tableposition=top]{caption}

\usepackage{algorithm}
\usepackage{algpseudocode}
\usepackage[dvipsnames]{xcolor}
\usepackage{color, colortbl}
\definecolor{Gray}{gray}{0.9}

\DeclareMathOperator*{\argmin}{arg\,min}
\def\wacvPaperID{201} % Enter the WACV Paper ID here

\wacvfinalcopy % *** Uncomment this line for the final submission

\ifwacvfinal
\def\assignedStartPage{9876} % *** Enter the assigned starting page number (instead of 9876)
\fi

\ifwacvfinal
\usepackage[breaklinks=true,bookmarks=false]{hyperref}
\else
\usepackage[pagebackref=true,breaklinks=true,colorlinks,bookmarks=false]{hyperref}
\fi

\ifwacvfinal
\else
\fi

\begin{document}

%%%%%%%%% TITLE
\title{Labeling Where Adapting Fails: Cross-Domain Semantic Segmentation with Point Supervision via Active Selection}

\author{Fei Pan\\
KAIST\\
{\tt\small feipan@kaist.ac.kr}
\and
Francois Rameau\\
KAIST \\
{\tt\small frameau@kaist.ac.kr}

\and
Junsik Kim\\
Harvard University\\
{\tt\small mibastro@gmail.com}

\and
In So Kweon\\
KAIST\\
{\tt\small iskweon77@kaist.ac.kr}
}

\maketitle
%\thispagestyle{empty}

%%%%%%%%% ABSTRACT
\begin{abstract}
% \FR{Note: I went through all the text, I think the methodology part is crystal clear! it was a pleasure to read, thank you for the high quality work~. and the figure is very cool too ;-). And plz add references where I put XX }
% Learning semantic segmentation models requires a large amount of pixel-wise annotated data, which might not exist in the desired target domain. Unsupervised domain adaptation approaches aim at aligning features of labeled source data and unlabeled target data. After all, these approaches produce limited performance due to the large domain gap. To cope with this limitation, previous works involve human annotators to provide labels for target image points which are randomly generated. In this work, we propose a new domain adaptation framework for semantic segmentation with annotated points via active selection. First, we conduct unsupervised domain adaptation of the model; from this adaptation, we use an entropy-based uncertainty measurement for target points selection. Finally, to minimize the domain gap, we propose a domain adaptation framework utilizing these target points annotated by human annotators. Experimental results on benchmark datasets show the effectiveness of our methods against existing unsupervised domain adaptation approaches.
Training models dedicated to semantic segmentation requires a large amount of pixel-wise annotated data. Due to their costly nature, these annotations might not be available for the task at hand. To alleviate this problem, unsupervised domain adaptation approaches aim at aligning the feature distributions between the labeled source and the unlabeled target data. While these strategies lead to noticeable improvements, their effectiveness remains limited. %when the domain gap is large \CN{why mention large gap here? This paper does not solve the problem of large gap. Maybe the motivation should be stated to reduce the annotation cost or achieving the balance of performance and annotation. I am not sure whether I am too biased. The performance improvement by th eactive learning strategy is too limited, thus I suggest focusing on the sparse annotation while active learning is just a free and cheap technique to boost the performance.}. \FR{Good point, we have to be careful (especially since we partly rely on UDA), not sure yet how to state it but we have to keep it in mind!! and we mention it a few times :s my bad, Maybe we can just say that "their effectiveness remains limited" ? without mention to gap? I think it is okay?}
To guide the domain adaptation task more efficiently, previous works attempted to include human interactions in this process under the form of sparse single-pixel annotations in the target data. In this work, we propose a new domain adaptation framework for semantic segmentation with annotated points via active selection.
First, we conduct an unsupervised domain adaptation of the model; from this adaptation, we use an entropy-based uncertainty measurement for target points selection.
Finally, to minimize the domain gap, we propose a domain adaptation framework utilizing these target points annotated by human annotators. Experimental results on benchmark datasets show the effectiveness of our methods against existing unsupervised domain adaptation approaches.
The propose pipeline is generic and can be included as an extra module to existing domain adaptation strategies.
\end{abstract}
%%%%%%%%% BODY TEXT
\section{Introduction}
\begin{figure}[t]
    \centering
    \includegraphics[width=0.45\textwidth]{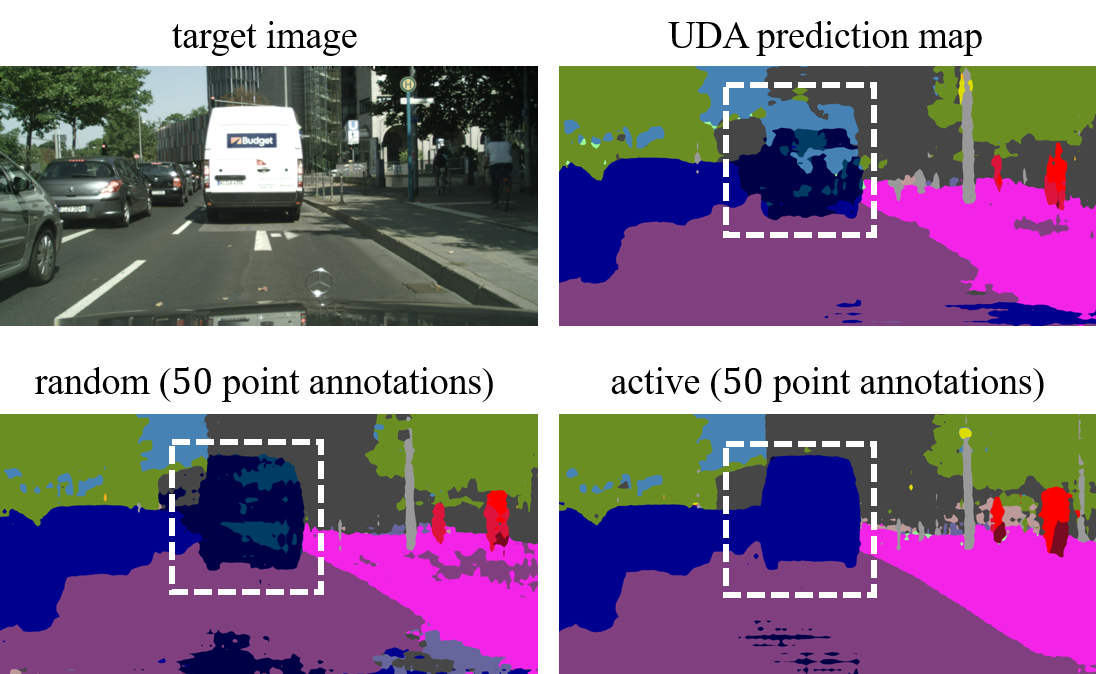}
    \caption{Existing UDA approaches' performance remains limited when the domain gap is big. We propose a domain adaptation framework for semantic segmentation with annotated target points via active selection. Under same labeling budget ($50$ point annotations), our approach utilizing point annotations via active selection outperforms the random selection method.}
    \label{fig:introduction}
\end{figure}

Semantic segmentation consists in assigning every image's pixel with its corresponding semantic class. With the recent improvements brought by convolution neural networks~\cite{long2015fully, chen2017deeplab, zhao2017pyramid}, semantic segmentation finds applications in a wide spectrum of tasks, such as, autonomous driving~\cite{luc2017predicting, Zhang_2017_ICCV}, robotics~\cite{milioto2018real} and disease diagnostic~\cite{zhou2019collaborative, zhao2019data}. Training semantic segmentation model is a data hungry process where a large number of pixel-wise annotated images is required. Typically, these annotations are manually assigned by human operators, which makes this task cumbersome, time consuming and, as a result, particularly expensive~\cite{Cordts2016Cityscapes}. While various computer applications have been successfully resolved in a self-supervised manner (such as, visual odometry~\cite{zhou2017unsupervised} or depth estimation~\cite{godard2017unsupervised}), semantic segmentation still requires to be trained in a supervised fashion. Despite the availability of large scale annotated datasets for semantic segmentation~\cite{Cordts2016Cityscapes, yu2020bdd100k, geiger2012we}, a network trained with this data is not guaranteed to generalize to a different set of images. This phenomenon is caused by the the distinct distributions between the two datasets, this discrepancy is known as \emph{domain gap}~\cite{ganin2015unsupervised}. To alleviate this problem, unsupervised domain adaptation (UDA) strategies have been proposed to align the feature distributions between the labeled source data and the unlabeled target data~\cite{pan2020unsupervised, inkyshin2020, park2020discover, hoffman2018cycada}. Specifically, adversarial learning-based UDA approaches has shown efficiency in aligning the domain gap at image~\cite{hoffman2018cycada}, features~\cite{saito2018maximum} or output level~\cite{tsai2018learning, tsai2019domain}. While UDA approaches lead to noticeable improvements, their effectiveness remains limited. % when the domain gap is large \CN{again, this motivation is flawed, unless the proposed technique is specifically designed to solve the "large" domain gap}. 

% As an essential task in computer vision, semantic segmentation aims at assigning each pixel in the image a semantic class. Recently, convolutional neural network-based segmentation models have achieved notable improvement, leading to their applications in multiple applications in computer vision systems, such as autonomous driving, robotics, and disease diagnosis. Training models dedicated to semantic segmentation requires a large amount of data with pixel-wise annotations. However, collecting these annotations for semantic segmentation task is difficult since they are costly and labor intensive. To alleviate this problem, unsupervised domain adaptation (UDA) for semantic segmentation approaches have been proposed to align the feature distributions between the labeled source data and the unlabeled target data. Specifically, adversarial learning-based UDA approaches has shown efficiency in aligning the domain gap at image, features or output level. Moreover, \cite{vu2019advent} propose to utilize the entropy of pixel-wise output predictions for output level alignment. 
% Other UDA approaches include generating pseudo labels for target data and retraining models with an iterative process. 
% While UDA approaches lead to noticeable improvements, their effectiveness remains limited when the domain gap is large. 

This remaining performance gap significantly diminish the relevance and practicability of these techniques for real-world applications where a high level of accuracy is required. To reduce the domain gap further, domain adaptation coupled with weak human annotation have been developed~\cite{Paul_WeakSegDA_ECCV20,wang2019weakly}. Instead of performing a full pixel-wise annotations of the target images, weak labeling consists in simpler and faster tasks, for instance, bounding box selection~\cite{dai2015boxsup, papandreou2015weakly}, image-level~\cite{ahn2018learning, chang2020weakly, kolesnikov2016seed} or points-level~\cite{bearman2016s} annotations. The cost of these weak annotations is significantly lower than their dense counterpart, making it realistically deplorable for industrial and commercial purpose. However, we argue that existing annotation process is yet to be optimized to achieve better performance under same labeling budget.

% To settle this issue, domain adaptation approaches for semantic segmentation with target weak annotations approaches \cite{Paul_WeakSegDA_ECCV20,wang2019weakly} have been developed. Under the labeling budget, these approaches directly label target data with bound box, image-level or point-level annotations. However, we argue that existing labeling process is still yet to be optimized to achieve better performance under same labeling budget.

In this work, we propose a new domain adaptation framework for semantic segmentation with annotated points via active selection. Our framework consists of three parts, which are presented in \figurename~\ref{fig:architecture}, namely, 1) an unsupervised domain adaptation to train a segmentation network and generates the entropy maps and predicted segmentation maps for all target training images, 2) an entropy-based label acquisition system to request annotations for points with high uncertainty level from oracle, while transferring UDA predicted labels to pseudo labels for points with low uncertainty level, 3) a new domain adaptation framework with target weak annotations to further align the domain shift. Our proposed approach achieves good performance against state-of-the-art UDA approaches on benchmark datasets. \\
%\CN{highlight that active learning just performs ordering, which is simple yet free}

\noindent \textbf{The Contributions of Our Work.} First, we introduce an entropy-based label acquisition method which is optimized to achieve better segmentation performance under equal labeling budget. Second, we propose a new domain adaptation model with target point-based weak annotations for semantic segmentation task. % \CN{I suggest switching the two points. point2 is actually the main framework, while point1 has to be based on point2, again free technique to improve the performance}. %I think it is okay (FR)

\section{Related Works}
In this section, we describe related methods for unsupervised domain adaptation and weakly supervised semantic segmentation. Moreover, we discuss about recent uncertainty estimation via entropy, which is a central component of the proposed technique.

\noindent \textbf{Unsupervised Domain Adaptation.} Unsupervised domain Adaptation (UDA) approaches aim at aligning the distribution shift between the labeled source data and the unlabeled target data. Recently, adversarial-based UDA approaches~\cite{park2020discover, pan2020unsupervised,  vu2019advent, tsai2019domain, hoffman2018cycada, tsai2018learning} have demonstrated to be effective in learning domain invariant features for semantic segmentation task. Adversarial-based UDA approaches for semantic segmentation contain two networks, one network is used as a generator to predict the segmentation maps of input images, which is given from either the source domain or target domain. Based on this, the second network, as a discriminator, predicts the domain labels for the features given from the generator. Then the generator tries to fool the discriminator, so as to align the distribution shift of features from the two domains. Besides alignment on feature level, other approaches propose to align domain shift at the image level~\cite{hoffman2018cycada}, output level~\cite{tsai2018learning} or entropy level~\cite{vu2019advent}. More recently,~\cite{tsai2019domain} proposes to align the domain shift by utilizing the patch-wise output distribution from the two domains. While existing UDA approaches lead to noticeable improvements, their effectiveness remains limited when the domain shift is large. In this paper, we combine a UDA approach with small labeling budget, \ie point annotation, to align domain gap effectively. \\

\noindent \textbf{Weakly Supervised Semantic Segmentation.} Weakly supervised semantic segmentation approaches utilize various types of weak semantic labels including image-level labels~\cite{ahn2018learning, chang2020weakly, kolesnikov2016seed}, bounding box~\cite{dai2015boxsup, papandreou2015weakly}, scribble labels~\cite{lin2016scribblesup}, and point labels~\cite{bearman2016s}. These approaches propose to train a segmentation model with these weak labels and test it in the same domain. Based on their frameworks, \cite{wang2019weakly} proposes to combine target bounding box annotation with domain adaptation to improve semantic segmentation performance from the source domain to the target domain. \cite{Paul_WeakSegDA_ECCV20} takes advantage of target weak point annotations with domain adaptation for better domain alignment. While existing approaches utilize target weak labels to reduce labeling cost in domain adaptation for semantic segmentation task, they fail to consider an optimized weak labeling policy. In this work, we propose a domain adaptation framework with point supervision via active selection to make the annotation process even more efficient. \\

\noindent \textbf{Uncertainty via Entropy.} Entropy as a measurement of uncertainty has been used in domain adaptation. \cite{vu2019advent} takes advantage of entropy of pixel-wise segmentation output to align the domain shift. Based on this, \cite{pan2020unsupervised} proposes entropy-based ranking system to split the target domain into two distinct parts: the easy and hard split. On this basis, an inter-domain and intra-domain adaptation are conducted to minimize the domain shift. Recently, \cite{su2020active} adopts the entropy of output as a confidence measurement for transferring samples across domains. Entropy functions as a uncertainty measurement has also been used widely in active learning for image recognition~\cite{luo2013latent,settles2008analysis,joshi2009multi}. There are a few works proposed recently applying active learning for semantic segmentation. \cite{kasarla2019region} applies entropy-based uncertainty measurement on super-pixels to find important regions to be labelled, then apply a conditional random field to propagate labels. Similarly, \cite{siddiqui2020viewal} proposes to use a view-point entropy combined with super-pixels to select informative samples for multi-view semantic segmentation. \cite{mackowiak2018cereals} use vote entropy~\cite{dagan1995committee} from dropout layers of segmentation model by constructing a Monte-Carlo dropout ensemble~\cite{gal2016dropout}.
While a number of works dedicated their effort to the efficient annotation for semantic segmentation, their scenarios are limited to a single domain. 

The effort to reduce the labeling budget has been researched in different directions separately: UDA, use of weak labels and active sampling. Although they aim for the same goal, \ie, reducing labeling cost, an attempt to harmonize these approaches has not been explored yet. We propose an active sampling of weak labels under a domain adaptation scenario suggesting a new important research direction.

% This work only considers data collected from a single domain, which is different from our work. In contrast, we propose to use an entropy-based point selection strategy in domain adaptation frameworks. 
% \CN{This part should highlight how active learning used in this paper is different from other works. For example, in normal active learning, it is used to choose the sample while this one chooses the patch, blabla, Contrary prior works blabla, the active learning technique adoped in this work blabla}
 
%%%%%%%%% BODY TEXT
\section{Methodology}
\begin{figure*}[t!]
    \centering 
    \includegraphics[width=0.97\textwidth]{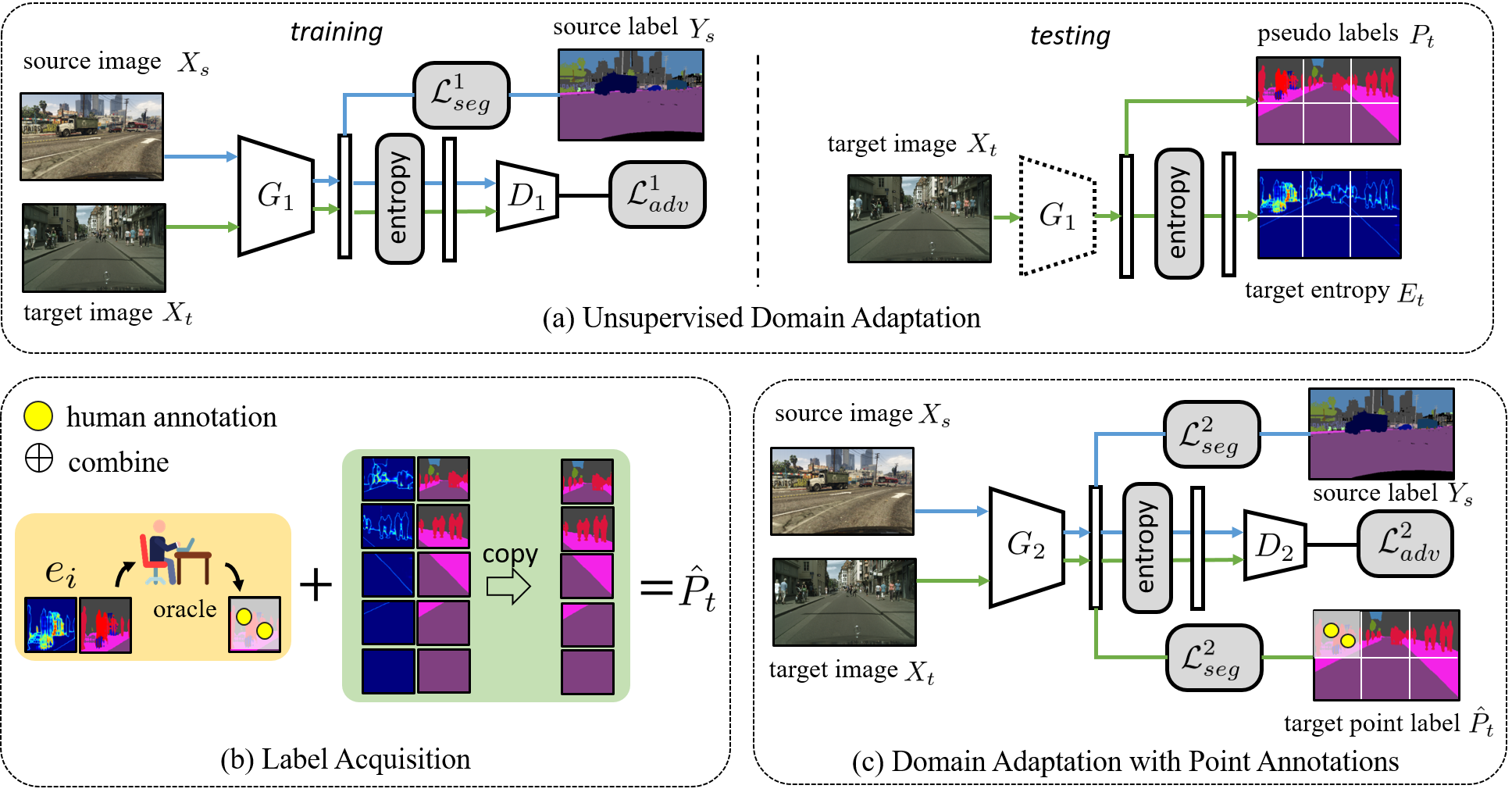}
    \caption{The proposed cross-domain semantic segmentation approach using target point annotations via active selection consists of three parts, namely, (a) an unsupervised domain adaptation (b) a label acquisition system, and (c) a domain adaptation using target point annotations. In (a), given the labeled source data and unlabeled target data, a segmentation model ~\cite{chen2017deeplab} $G_1$ is trained to generate prediction maps as pseudo labels and entropy maps; $D_1$ is trained to predict the domain label for the entropy maps while $G_1$ is trained to fool $D_1$. $\{G_1, D_1\}$ are optimized by minimizing the segmentation loss $\mathcal{L}_{seg}^1$ and the adversarial loss $\mathcal{L}_{adv}^1$. In (b), an entropy-based label acquisition system is used to locate the most uncertain predictions in the target image, on which a human annotation is pertinent. The \textcolor{Dandelion}{yellow} dots represents locations of the selected points for human annotation, while in other patches the pseudo labels are copied and preserved. In (c), a domain adaption method with point annotations from target domain is proposed. Given the data from both domains, the semantic segmentation model $G_2$ is trained to generate predictions maps and entropy maps. $D_2$ is used to predict whether the samples is from the source domain or the target domain, while $G_2$ is used to fool $D_2$. $\{G_2, D_2\}$ is optimized using a cross-entropy based segmentation loss $\mathcal{L}_{seg}^2$ and adversarial loss $\mathcal{L}_{adv}^2$. }
    \label{fig:architecture}
\end{figure*}

In this section, a cross-domain semantic segmentation approach using target point annotations via an active selection is introduced. The proposed approach consists of three parts. First, a cross-domain segmentation model~\cite{chen2017deeplab} $G_1$ is trained using the labeled source data and the unlabeled target data. The model $G_1$ takes each target data to generate prediction maps and entropy maps. 
%Secondly, an entropy-based uncertainty measurement is used to select patches of target data and request for point annotation from oracles. 
Secondly, an entropy-based uncertainty measurement is used to automatically select ambiguous patches in the target images and request for point annotation from oracles. 
Lastly, a simple yet effective domain adaptation method using target weak and pseudo-label annotations to train a second semantic segmentation model~\cite{chen2017deeplab} $G_2$ is proposed.

Let $\mathcal{S}$ denote a source domain containing a set of RGB images and their corresponding pixel-wise annotations $\{ X_s^i, Y_s^i \}_{s=1}^{N_s}$, where $X_s^i \in \mathbb{R}^{H \times W \times 3}$, $Y_s^i \in \mathbb{R}^{H \times W \times C}$; where $N_s$ and $C$ stands for to the number of image-label pairs and the number of classes (for both source domain and target domain), respectively. 
Similarly, let $\mathcal{T}$ denote a target domain containing a set of unlabelled images $\{ X_t^i\}_{t=1}^{N_t}$, where $X_t^i \in \mathbb{R}^{H \times W \times 3}$ and $N_t$ is the number of target data.

% Let $\mathcal{S}$ denote a source domain containing a set of RGB images and \FR{their corresponding} pixel-wise annotations $\{ X_s^i, Y_s^i \}_{s=1}^{N_s}$, where $X_s^i \in \mathbb{R}^{H \times W \times 3}$, $Y_s^i \in \mathbb{R}^{H \times W \times C}$, and $N_s$ is the number of source data; let $\mathcal{T}$ denote a target domain containing a set of unlabelled images $\{ X_t^i\}_{t=1}^{N_t}$, where $X_t^i \in \mathbb{R}^{H \times W \times 3}$ and $N_t$ is the number of target data. Note that $C$ denotes the number of categories for both source domain and target domain. 

\subsection{Unsupervised Domain Adaptation}
It is assumed that the target domain exclusively contains unlabeled RGB images before any request for annotations from oracle. In order to estimate the adaptation uncertainty on the target data, an unsupervised domain adaptation (UDA) is conducted using a semantic segmentation model $G_1$. Given a sample $X_s$ from the source domain alongside with its ground-truth annotation $Y_s$, the network $G_1$ takes $X_s$ as an input and generates a ``soft-segmentation map" $P_s=G_1(X_s)$. In this map, each $C$ dimensional vector $\left[ P_s^{(h,w,c)} \right]_C$ at a pixel position $(h,w)$ is a discrete probability distribution over $C$ classes. The segmentation model $G_1$ is optimized through a cross-entropy loss:
\begin{equation}
    \mathcal{L}_{seg}^{1}(X_s, Y_s) = - \sum_{h,w,c} Y_s^{(h,w,c)}\log P_s^{(h,w,c)}.
\end{equation} 
To bridge the domain gap between the source and the target domain, ADVENT~\cite{vu2019advent} aligns the entropy maps from the source domain and the target domain by using an adversarial training technique. The assumption of~\cite{vu2019advent} is that the trained models tend to produce high-confident (low-entropy) predictions on source-like data, and low-confident (high-entropy) predictions on target-like data. Given an input $X_t$ from the target domain, the segmentation model $G_1$ produces the segmentation map $P_t=G_1(X_t)$. The entropy map $E \in \mathbb{R}^{H \times W \times C}$ is formulated as: 
\begin{equation}
\label{eq-2}
E_t^{(h,w,c)} = -P_t^{(h,w,c)} \log P_t^{(h,w,c)}.
\end{equation}
In order to align features from the source domain and the target domain, a domain discriminator $D_1$ is trained to produce a domain label given the entropy map as an input. In other words, $D_1$ produces $1$ when the entropy map is from source domain, and $0$ when it is from the target domain. On this basis, $G_1$ is trained to fool $D_1$. The optimization of $G_1$ and $D_1$ is achieved by the adversarial loss function:
\begin{equation}
    \begin{split}
            \mathcal{L}_{adv}^{1} (X_s, X_t) =  \sum_{h,w,c}  \log [ 1 - D_{1}(E_t^ {(h,w,c)})] & \\
                                    +     \log D_{1}(E_s^{(h,w,c)}) &, 
    \end{split}
\end{equation}
where $E_s$ is the entropy map of $X_s$. The complete loss function of unsupervised domain adaptive segmentation is formulated as:
\begin{equation}
    \label{eq-4}
    \begin{split}
    \min_{G_1} \max_{D_1}  \mathcal{L}_{seg}^{1}(X_s,Y_s) + \mathcal{L}_{adv}^{1}(X_s,X_t).
    \end{split}
\end{equation}

\subsection{Label Acquisition}
Due to a large domain gap, the segmentation model $G_1$ trained on labeled source data and unlabeled target data might still lead to poor performance on the target domain. In order to reduce the domain gap further, we propose to use additional information from human annotators on the target data. For monetary reasons~\cite{bearman2016s}, instead of relying on dense annotations, oracles are requested to provide point-level weak annotations on the target data. Concretely, an target image containing a single landmark point is provided to the annotator who has to select the class corresponding to this pixel. While the location of these points is sometimes assigned randomly~\cite{Paul_WeakSegDA_ECCV20}. In this work, we designed an entropy-based selection method to automatically select the most informative points to be annotated. Once the segmentation model $G_1$ is optimized via \eqref{eq-4}, given a target image $X_t$ as an input, $G_1$ generates its segmentation map $P_t$ and entropy map $E_t$. This entropy map is used to locate the most uncertain predictions in the image $X_t$ on which a human annotation would be pertinent. Specifically, $E_t$ is divided by a $M \times N$ regular grid into multiple patches $\{ e_t^i\}_{i=1}^{M\times N}$, where $E_t=\{e_t^i\}_{i=1}^{M\times N}$. Note that $e_t^i \in \mathbb{R}^{H' \times W' \times C}$, where $H'$ and $W'$ are the height and width of patches. 
In order to localize the uncertainty prediction area on target data, we take consideration of the mean entropy $e^{i}$ of the patch, which is given by:

% Due to a large domain gap, the segmentation model $G_1$ trained on labeled source data and unlabeled target data might still lead poor performance on target domain. 
% The reason behind is that pixel-wise annotations for target images are practically difficult to obtain. However, providing point-level weak annotations of target data is comparably cheap in semantic segmentation task~\cite{bearman2016s}. In this part, an entropy-based selection method is introduced to select and annotate most informative points of target data. Once the segmentation model $G_1$ is optimized via \eqref{eq-4}, given a target data $X_t$ as an input, $G_1$ generates its segmentation map $P_t$ and entropy map $E_t$. To estimate the uncertainty of $G_1$ on which area of $X_t$, we take advantage of $E_t$ to determine the confidence level of target predictions. Specifically, $E_t$ is divided by a grid of $M \times N$ into multiple patches $\{ e^i\}_{i=1}^{M\times N}$, where $E_t=\{e^i\}_{i=1}^{M\times N}$. Note that $e^i \in \mathbb{R}^{H' \times W' \times C}$, where $H'$ and $W'$ are the height and width of patches. In order to localize the uncertainty prediction area on target data, we take consideration of the mean entropy $e^{i}$ of the patch, which is given by:
\begin{equation}
    \label{eq-5}
    \epsilon^i = \sum_{h',w',c} {e_t^i}^{(h', w', c)}.
\end{equation}

As an uncertainty measurement, $\epsilon^i$ with a high value suggests that the model $G_1$ produce high uncertainty predictions in the cell $e_t^i$. Thus, an oracle is requested to label one (or multiple) points in the most uncertain patches of the image. 
Given a selected patch $e_t^i$, a random point position $\{a, b\}$ is chosen inside the patch;
If $\epsilon^i$ is among the top $K$ highest mean entropy among all patches, then a request is sent for oracles to label the point positions $\{a,b\}$; otherwise, the pseudo label from the predicted map $P_t$ (from UDA) will be transferred at $\{a, b\}$ as this prediction is assumed to be confident. While for all the other non-selected patches, we directly copy its pseudo labels, as shown in the green area of ~\ref{fig:architecture} (b). Let $\hat{P}_t$ denote a new label, the annotation process is formulated as follows:

% As an uncertainty measurement, $\epsilon^i$ with a high value
% suggests that the model $G_1$ produce high uncertainty predictions at the positions covered within the patch $e^i$. Thus, a feasible idea is to request annotations for high mean entropy patches from oracle.
% Given a patch $e^i$, a random point position $\{a, b\}$ is generated from the patch, and the mean entropy $\epsilon^i$ is calculated through \eqref{eq-5}. If $\epsilon^i$ is among the top $K$ highest mean entropy for all the patches, then a request is sent for oracles to label point positions $\{a,b\}$; otherwise, the a pseudo label will be transferred at $\{a, b\}$ from the predicted map $P_t$ generated by UDA. Let $\hat{P}_t$ denote a new label, the labeling process is formulated as follows:
\begin{equation}
    \hat{P}_t(e_t^i)= 
    \begin{cases}
        \text{oracle}, & \text{ if } \epsilon_t^i \in {\text{top $K$ highest}}\\
        P_t(e_t^i), & \text{otherwise}     
    \end{cases} . 
\end{equation}
% \FR{I added the dot ;-), And the "otherwise" was missing, I used this: https://tinyurl.com/y3a2oebo }

%-----------------------------------------GTA 5--------------------------------

\begin{table*}[ht!]
\centering
\caption{The semantic segmentation results of Cityscapes validation set with trained models adapted from GTA5 to Cityscapes. All the results are generated from the ResNet-101-based models. IntraDA~\cite{pan2020unsupervised} is used as our UDA baseline model for the experiments. Self-training denotes the performance generated from experiments by only using pseudo labels from first stage of IntraDA. Random selection denotes the performance generated from experiments by randomly selecting 10 patches with each 5 human point annotations are provided. Active selection denotes the performance generated from experiments by selecting top 10 highest entropy patches with each 5 human point annotations are provided.  Full labeling denotes the performance generated from experiments by selecting top 10 highest patches with each full human point annotations are provided. mIoU denotes the mean IoU over 19 classes.}
\resizebox{\textwidth}{!}
{
\begin{tabular}{l|c c c c c c c c c c c c c c c c c c c|c}
\toprule[1.0pt]
\multicolumn{21}{c}{ GTA5 $\to$ Cityscapes}\\
% \hline \hline
\toprule[1.0pt]
Method & \rotatebox{90}{road} & \rotatebox{90}{sidewalk} & \rotatebox{90}{building} & \rotatebox{90}{wall} & \rotatebox{90}{fence} & \rotatebox{90}{pole} & \rotatebox{90}{light} & \rotatebox{90}{sign} & \rotatebox{90}{veg} & \rotatebox{90}{terrain} & \rotatebox{90}{sky} & \rotatebox{90}{person} & \rotatebox{90}{rider} & \rotatebox{90}{car}& \rotatebox{90}{truck} & \rotatebox{90}{bus} & \rotatebox{90}{train} & \rotatebox{90}{mbike} & \rotatebox{90}{bike} & mIoU \\
\hline
No Adaptation~\cite{tsai2018learning}  & 75.8 & 16.8 & 77.2 & 12.5 & 21.0 & 25.5 & 30.1 & 20.1 & 81.3 & 24.6 & 70.3 & 53.8 & 26.4 & 49.9 & 17.2 & 25.9 & 6.5 & 25.3 & 36.0 & 36.6 \\
ROAD~\cite{chen2018road}                    & 76.3 & 36.1 & 69.6 & 28.6 & 22.4 & 28.6 & 29.3 & 14.8 & 82.3 & 35.3 & 72.9 & 54.4 & 17.8 & 78.9 & 27.7 & 30.3 & 4.0 & 24.9 & 12.6 & 39.4 \\
AdaptSegNet~\cite{tsai2018learning}         & 86.5 & 36.0 & 79.9 & 23.4 & 23.3 & 23.9 & 35.2 & 14.8 & 83.4 & 33.3 & 75.6 & 58.5 & 27.6 & 73.7 & 32.5 & 35.4 & 3.9 & 30.1 & 28.1 & 42.4 \\
CLA~\cite{luo2019taking}                    & 87.0 & 27.1 & 79.6 & 27.3 & 23.3 & 28.3 & 35.5 & 24.2 & 83.6 & 27.4 & 74.2 & 58.6 & 28.0 & 76.2 & 33.1 & 36.7 & 6.7 & 31.9 & 31.4 & 43.2 \\
MinEnt~\cite{vu2019advent}                  & 84.2 & 25.2 & 77.0 & 17.0 & 23.3 & 24.2 & 33.3 & 26.4 & 80.7 & 32.1 & 78.7 & 57.5 & {30.0} & 77.0 & {37.9} & 44.3 & 1.8 & 31.4 & 36.9 & 43.1 \\
AdvEnt~\cite{vu2019advent}                  & 89.9 & 36.5 & 81.6 & 29.2 & 25.2 & 28.5 & 32.3 & 22.4 & 83.9 & 34.0 & 77.1 & 57.4 & 27.9 & 83.7 & 29.4 & 39.1 & 1.5 & 28.4 & 23.3 & 43.8 \\
SWD~\cite{lee2019sliced}                  & 92.0 & 46.4 & 82.4 & 24.8 & 24.0 & 35.1 & 33.4 & 34.2 & 83.6 & 30.4 & 80.9 & 56.9 & 21.9 & 82.0 & 24.4 & 28.7 & 6.1 & 25.0 & 33.6 & 44.5  \\
SSF-DA~\cite{du2019ssf}                   &  90.3 & 38.9 & 81.7 & 24.8 & 22.9 & 30.5 & 37.0 & 21.2 & 84.8 & 38.8 & 76.9 & 58.8 & 30.7 & 85.7 & 30.6 & 38.1 & 5.9 & 28.3 & 36.9  & 45.4 \\
DISE~\cite{chang2019all}                  & 91.5 & 47.5 & 82.5 & 31.3 & 25.6 & 33.0 & 33.7 & 25.8 & 82.7 & 28.8 & 82.7 & 62.4 & 30.8 & 85.2 & 27.7 & 34.5 & 6.4 & 25.2 & 24.4 & 45.4 \\
IntraDA~\cite{pan2020unsupervised} & 90.6 & 36.1 & 82.6 & 29.5 & 21.3 & 27.6 & 31.4 & 23.1 &85.2 & 39.3 & 80.2 &59.3 & 29.4 & 86.4 & 33.6 & 53.9 & 0.0 & 32.7 & 37.6 & 46.3 \\ 
AdaptPatch~\cite{tsai2019domain}          & 92.3 & 51.9 & 82.1 & 29.2  & 25.1 & 24.5 & 33.8 & 33.0 & 82.4 & 32.8 & 82.2  & 58.6  & 27.2  & 84.3 & 33.4 & 46.3 & 2.2 & 29.5 & 32.3 & 46.5\\
\hline
Self-training (lower bound)                             & 91.8 & 50.1 & 84.7 & 33.3 & 28.1 & 30.3 & 36.9 & 29.1 & 84.1 & 34.2 & 86.4 & 60.1 & 31.8 & 83.1 & 25.5 & 46.7 & 0.0 & 28.1 & 40.2 & 47.7 \\
Ours - Random Selection           & 92.2 & 51.7 & 84.8 & 32.8 & 28.2 & 31.0 & 37.6 & 31.2 & 84.2 & 35.1 & 85.3 & 61.7 & 32.3 & 83.2 & 29.4 & 49.3 & 0.2 & 33.5 & 42.3 & 48.7 \\ 
Ours - Active Selection           & 92.7 & 55.6 & 85.4 & 35.9 & 30.5 & 33.7 & 40.2 & 35.0 & 85.2 & 38.9 & 86.3 & 63.9 & 36.5 & 84.6 & 30.7 & 49.3 & 1.8 & 38.2 & 54.7 & 51.5\\
\rowcolor{gray!40}
Ours - Full Labeling (upper bound)                             & 94.9 & 68.4 & 86.2 & 42.3 & 36.1 & 37.7 & 39.8 & 52.4 & 86.9 & 47.6 & 88.6 & 64.6 & 40.4 & 88.6 & 59.8 & 65.5 & 40.0 & 40.3 & 61.1 & \textbf{60.1} \\
\bottomrule
\end{tabular}}
\label{tab:gta5}
\end{table*}
% -------------------------------------------------------------------

\subsection{Domain Adaptation with Weak Annotations}

In this section, a domain adaptation method using target weak annotations $\hat{P}_t$ to train a semantic segmentation network $G_2$ is introduced. In order adapt the segmentation model $G_2$ into the target domain, we use target weak annotations to optimize $G_2$ via a cross-entropy loss. Given a sample $X_s$ from the source domain and $X_t$ from the target domain, we forward them into $G_2$ and obtain the segmentation maps $P_s'=G_2(X_s)$ and $P_t'=G_2(X_t)$. With the aid of the source annotation $Y_s$ and the target weak annotation $\mathcal{Y}_t$, the segmentation model $G_2$ is optimized by the segmentation loss:
\begin{equation}
    \begin{split}
            \mathcal{L}_{seg}^{2} (X_s, Y_s, X_t,\hat{P}_t) = - \sum_{h,w,c} Y_s^{(h,w,c)}  \log G_2(X_s^{(h,w,c)})  \\
                               + \hat{P}_t^{(h,w,c)}\log G_2(X_t^{(h,w,c)}) &.
    \end{split}
\end{equation}
To bridge the gap between the source domain and the target domain, we adopt an entropy-based alignment strategy for both domains. Given the predicted maps $P_s'$ and $P_t'$, we calculate the corresponding entropy maps $E_s'$ and $E_t'$ by using \eqref{eq-2}. To close the domain gap, a discriminator $D_2$ is trained to predict the domain labels of $E_s'$ and $E_t'$. And $G_2$ is trained to fool $D_2$. The adversarial training loss to optimize $G_2$ and $D_2$ is given by:
\begin{equation}
    \begin{split}
            \mathcal{L}_{adv}^{2} (X_s, X_t) =  \sum_{h,w,c}  \log [ 1 - D_{1}(E_t'^ {(h,w,c)})] & \\
                                    +     \log D_{1}(E_s'^{(h,w,c)}) &.
    \end{split}
\end{equation}
In the end, our complete loss function is formed by:
\begin{equation}
    \mathcal{L} = \mathcal{L}_{seg}^{2} (X_s, Y_s, X_t, \hat{P}_t) +  \mathcal{L}_{adv}^{2} (X_s, X_t).
\end{equation}
Our goal is to learn a target segmentation model ${G_2}^{*}$ through:
\begin{equation}
    G_2^{*} = \argmin_{G_2} \min_{G_2} \max_{D_2}  \mathcal{L}.
\end{equation}

% \begin{algorithm}
% 	\caption{My Algorithm}
% 	\begin{algorithmic}[1]
% 	\State Given $P_t$, $\{e_t^i\}_{i=1}^{M\times N}$, $K$, $N$
% 		\For {$i=1,2,\ldots$}
% 		\If{$\epsilon_t^i$ is among the top $K$ highest entropy value}
% 		    \State choose random
% 		\EndIf
% 			\For {$actor=1,2,\ldots,N$}
% 				\State Run policy $\pi_{\theta_{old}}$ in environment for $T$ time steps
% 				\State Compute advantage estimates $\hat{A}_{1},\ldots,\hat{A}_{T}$
% 			\EndFor
% 			\State Optimize surrogate $L$ wrt. $\theta$, with $K$ epochs and minibatch size $M\leq NT$
% 			\State $\theta_{old}\leftarrow\theta$
% 		\EndFor
% 	\end{algorithmic} 
% \end{algorithm}

% \begin{equation}
%     e_t^i \in 
%     \begin{cases}
%       \mathbb{E}_{or} ,& \text{if } \epsilon_t^i \text{ is among top } N \\
%         0,              & \text{otherwise}
%     \end{cases}
    
% \end{equation}

%-------------------------------------SYNTHIA ------------------------------------
\begin{table*}[ht!]
\centering
\caption{The semantic segmentation results of Cityscapes validation set with trained models adapted from SYNTHIA to Cityscapes. All the results are generated from the ResNet-101-based models. IntraDA~\cite{pan2020unsupervised} is used as our UDA baseline model for the experiments. Self-training denotes the performance generated from experiments by only using pseudo labels from first stage of IntraDA. Random selection denotes the performance generated from experiments by randomly selecting 10 patches with each 5 human point annotations are provided. Active selection denotes the performance generated from experiments by selecting top 10 highest entropy patches with each 5 human point annotations are provided.  Full labeling denotes the performance generated from experiments by selecting top 10 highest patches with each full human point annotations are provided. mIoU denotes the mean IoU over 19 classes. mIoU$^{*}$  denotes the mean IoU of 13 classes, excluding the classes with $^{*}$.}
\resizebox{\textwidth}{!}{
\begin{tabular}{l|c c c c c c c c c c c c c c c c|c|c}
\toprule[1.0pt]
\multicolumn{19}{c}{SYNTHIA $\to$ Cityscapes}\\
% \hline \hline
\toprule[1.0pt]
Method & \rotatebox{90}{road} & \rotatebox{90}{sidewalk} & \rotatebox{90}{building} & \rotatebox{90}{wall$^{*}$} & \rotatebox{90}{fence$^{*}$} & \rotatebox{90}{pole$^{*}$} & \rotatebox{90}{light} & \rotatebox{90}{sign} & \rotatebox{90}{veg} & \rotatebox{90}{sky} & \rotatebox{90}{person} & \rotatebox{90}{rider} & \rotatebox{90}{car}&  \rotatebox{90}{bus} & \rotatebox{90}{mbike} & \rotatebox{90}{bike} & mIoU & mIoU$^{*}$\\
\hline
No adaptation~\cite{tsai2018learning}       & 55.6 & 23.8 & 74.6 & 9.2  & 0.2 & 24.4 & 6.1 & {12.1} & 74.8 & 79.0 & 55.3 & 19.1 & 39.6 & 23.3 & 13.7 & 25.0 & 33.5 & 38.6 \\
AdaptSegNet~\cite{tsai2018learning}         & 81.7 & 39.1 & 78.4 & {11.1} & 0.3 & 25.8 & 6.8 & 9.0  & 79.1 & 80.8 & 54.8 & 21.0 & 66.8 & 34.7 & 13.8 & 29.9 & 39.6 & 45.8 \\
MinEnt~\cite{vu2019advent}                  & 73.5 & 29.2 & 77.1 &  7.7 & 0.2 & {27.0} & 7.1 & 11.4 & 76.7 & 82.1 & {57.2} & 21.3 & 69.4 & 29.2 & 12.9 & 27.9 & 38.1 & 44.2  \\
AdvEnt~\cite{vu2019advent}                  & {87.0} & {44.1} & {79.7} & 9.6 & {0.6} & 24.3 & 4.8 & 7.2 & {80.1} & 83.6 & 56.4 & {23.7} & 72.7 & 32.6 & 12.8 & 33.7 & 40.8 & 47.6  \\
CLAN~\cite{luo2019taking}                   & 81.3 & 37.0 & 80.1 & - & - & - & 16.1 & 13.7 & 78.2 & 81.5 & 53.4 & 21.2 &  73.0 & 32.9 & 22.6 & 30.7 & - & 47.8 \\
SWD~\cite{lee2019sliced}                    & 82.4 & 33.2 & 82.5 & - & - & - & 22.6 & 19.7 & 83.7 & 78.8 & 44.0 & 17.9 &  75.4 & 30.2 & 14.4 & 39.9 & - & 48.1 \\
DADA~\cite{vu2019dada}                      & 89.2 & 44.8 & 81.4 & 6.8 & 0.3 & 26.2 & 8.6 & 11.1 & 81.8 & 84.0 & 54.7 & 19.3 & 79.7 & 40.7 & 14.0 & 38.8 & 42.6 & 49.8 \\
SSF-DAN~\cite{du2019ssf}                    & 84.6 & 41.7 & 80.8 & - & - & - & 11.5 & 14.7 & 80.8 & 85.3 & 57.5 & 21.6 & 82.0 & 36.0 & 19.3 & 34.5 & - & 50.0 \\
DISE~\cite{chang2019all}                    & 91.7 & 53.5 & 77.1 & 2.5 & 0.2 & 27.1 & 6.2 & 7.6 & 78.4 & 81.2 & 55.8 & 19.2 & 82.3 & 30.3 & 17.1 & 34.3 & 41.5 & 48.8 \\
AdaptPatch~\cite{tsai2019domain}            & 82.4 & 38.0 & 78.6 & 8.7 & 0.6 & 26.0 & 3.9 & 11.1 & 75.5 & 84.6 & 53.5 & 21.6 & 71.4 & 32.6 & 19.3 & 31.7 & 40.0 & 46.5 \\
IntraDA~\cite{pan2020unsupervised}          & 84.3 & 37.7 & 79.5 & 5.3 & 0.4 & 24.9 & {9.2} & 8.4 & 80.0 & {84.1} & {57.2} & 23.0 & {78.0}& {38.1} & {20.3} & {36.5} & {41.7} & {48.9}   \\
\hline
Self-training (lower bound)                          & 85.2 & 37.3 & 80.8 & 5.2 & 0.4 & 24.4 & 9.0 & 10.3 & 81.3 & 83.5 & 56.4 & 22.6 & 78.3 & 37.0 & 22.1 & 37.4 & 42.0 & 49.3 \\
Ours - Random Selection              & 89.7 & 48.6 & 81.2 & 24.4 & 22.8 & 29.3 & 28.7 & 25.5 & 77.2 & 81.7 & 60.7 & 31.0 & 81.2 & 41.8 & 31.1 & 41.4 & 49.8 & 55.4 \\
Ours - Active Selection              & 91.9 & 54.9 & 84.7 & 34.2 & 25.5 & 33.9 & 31.7 & 32.5 & 84.6 & 87.1 & 63.9 & 37.5 & 85.5 & 52.0 & 36.6 & 54.3 & 55.8 & 61.3 \\
\rowcolor{gray!40} 
Ours - Full Labeling (upper bound)                              & 93.5 & 66.3 & 85.5 & 39.0 & 37.0 & 36.3 & 37.0 & 53.7 & 85.6 & 89.1 & 65.4 & 40.8 & 85.2 & 56.4 & 39.4 & 60.9 & 60.7 & \textbf{66.2} \\
\bottomrule
\end{tabular}}
\label{tab:synthia}
\end{table*}
% -------------------------------------------------------------------
\section{Experiments}
% In this section, we present the evaluation studies of the proposed domain adaptation framework for semantic segmentation with annotated points via active selection.

\subsection{Datasets}
To evaluate our proposed approach with state-of-the-art UDA approaches~\cite{tsai2018learning, vu2019advent, luo2019taking, lee2019sliced, pan2020unsupervised}, we adopt the same setting of adaptation from the synthetic to the real domain. To conduct this series of tests, synthetic datasets including GTA5~\cite{richter2016playing} and SYNTHIA~\cite{ros2016synthia} are used as source domains, along with the real-world dataset Cityscapes~\cite{Cordts2016Cityscapes} as the target domain. In the training period, the models are given labeled source data and unlabeled target data as inputs. After training, the models are evaluated on Cityscapes validation set. 

\begin{itemize}
    \item GTA5: we adopt the synthetic dataset GTA5~\cite{richter2016playing} which contains 24,966 synthetic images and corresponding ground-truth annotations of $1{,}914\times1{,}052$. All of these synthetic images are generated from a video game engine which mimics the urban scenery of Los Angeles city. And the ground-truth annotations are produced automatically by programs from this game engine. The dataset covers a total of 33 classes. However, in consideration of the compatibility with the Cityscapes dataset, we use 19 out of 33 classes: road, sidewalk, building, wall, fence, pole, traffic light, traffic sign, vegetation, terrain, sky, person, rider, car, truck, bus, train, motor bike, and bike.
    
    \item SYNTHIA: SYNTHIA-RAND-CITYSCAPES~\cite{ros2016synthia} as another synthetic dataset contains 9,400 fully annotated RGB images. Due to the large domain gap, number of samples of terrain, truck and train have limited distribution in the source domain. Thus 16 common classes in the experiments from SYNTHIA to Cityscapes dataset are considered in the experiment. In the process of testing, we take consideration of both 16 classes and 13 classes for better evaluation.
    
    \item Cityscapes: Cityscapes~\cite{Cordts2016Cityscapes} provides images collected from real urban scenes with fine segmentation annotations. In the training set, there are 2,975 images with full segmentation maps. However, this work does not use any annotation maps for the training images. Finally we evaluate our model on the validation set, which contains 500 images with annotations.
\end{itemize}

\noindent \textbf{Evaluation.} In this work,  we evaluate the semantic segmentation performance based on every category using the intersection-over-union metric, \ie, $\text{IoU} = {\text{TPO}}/{(\text{TPO}+\text{FPO}+\text{FNE})}$~\cite{everingham2015pascal}, where TPO, FPO, and FNE represent the number of true positive, false positive, and false negative pixels, respectively. \\

\begin{figure*}[ht!]
    \centering
    \includegraphics[width=0.94\textwidth]{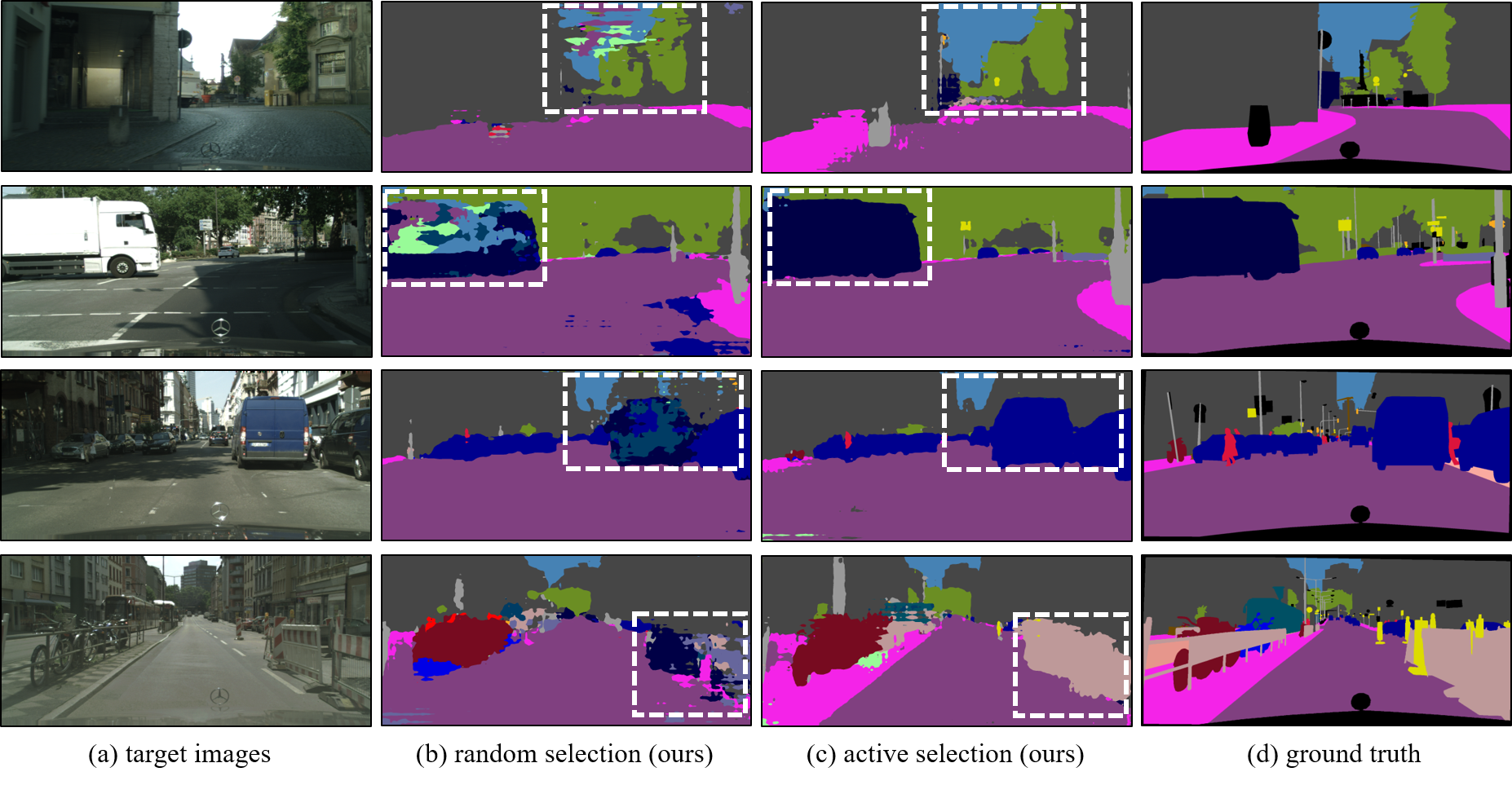}
    \caption{The qualitative results of evaluation for GTA5 $\to$ Cityscapes. (a) and (d) are the images and corresponding ground truth annotations from Cityscapes validation set. (b) are the prediction maps from the model using random selection strategy. (c) are the prediction maps from the proposed model using active selection. Note that these experiments are conducted with number of selected patches is 10, and in each patch 5 random points are selected for oracle to annotate.}
    \label{fig:qualitative_1}
\end{figure*}

\noindent \textbf{Implementation Details.} Due to its effectiveness in minimizing the domain shift, we adopt the framework of~\cite{pan2020unsupervised} for unsupervised domain adaptation, in both two experiments of GTA5$\to$Cityscapes and SYNTHIA$\to$Cityscapes. For the first stage of unsupervised domain adaptation, the backbone of $G_1$ is a ResNet-101 architecture~\cite{he2016deep} with pretrained parameters from ImageNet~\cite{deng2009imagenet}. With respect to the size of input data, we provide for the model $G_1$ the labeled source data of $1{,}280\times 760$ and unlabeled target data of $1{,}024\times 512$. The model $G_1$ is trained for totally 120,000 iterations. After training, $G_1$ is used to generate the corresponding prediction maps and entropy maps for all 2,975 images from Cityscapes training set. During the second stage of label acquisition, we divide each entropy map with a grid of $16\times 8$ into $128$ patches; the size of each patch is $64\times 64$. On this basis, we randomly generated $N$ points to be labeled. If the patch mean entropy is among top $K$ highest, we send a request for human annotator to label these points. Otherwise, we directly transfer the pseudo labels from the prediction maps (from UDA) to the selected points. For the third stage of domain adaptation with point annotations, we train a semantic segmentation model $G_2$ which has same architecture as $G_1$, and a discriminator $D_2$ similar to $D_1$; the input target data are 2,975 Cityscapes training images with target point label $\hat{P}_t$. $G_2$ is trained with the pretrained parameters from ImageNet and $D_2$ is trained from scratch.

Similarly to~\cite{pan2020unsupervised}, ~\cite{vu2019advent} and~\cite{tsai2018learning},  we utilize the multi-level feature outputs from \textit{conv4} and \textit{conv5} for both the first and the second stage of domain adaptation. To train $G_1$ and $G_2$, we apply an SGD optimizer~\cite{bottou2010large} with a learning rate of $2.5\times 10^{-4}$, momentum $0.9$, and a weight decay $10^{-4}$ for training $G_1$ and $G_2$. An Adam optimizer~\cite{kingma2014adam} with a learning rate of $10^{-4}$ is used for training $D_1$ and $D_2$.

%---------------------------hyperparameter K --------------------------------
\begin{table}[t!]
    \centering
    \caption{The ablation study on hyperparameter $K$ for patch selection. Note that the number of patches is $128$. Thus the maximum value of $K$ is $128$. We select patches with top $K$ highest mean entropy, and from each patch we randomly generate 5 points for oracle to annotate.}
    \resizebox{0.45\textwidth}{!}{
    \begin{tabular}{c|c c c c c c c}
    \toprule[1.0pt]
    \multicolumn{8}{c}{GTA5 $\to$ Cityscapes}\\
    \hline
    $K$  & 0    & 1     & 2     & 3     & 5     &    10 &   128 \\
    \hline
    mIoU & 47.7 & 49.2 & 49.3 & 49.5 & 50.1 & 51.5 & 60.1\\
    \bottomrule
    \end{tabular}}
    \label{tab:K}
\end{table}
%-------------------------------------------------------------------------------
\subsection{Results}

\noindent \textbf{GTA5.} We compare the semantic segmentation performance of our propose approach against state-of-the art UDA approaches on Cityscapes Validation set, as shown in Table~\ref{tab:gta5} and Table~\ref{tab:synthia}. For a fair comparison, the baseline is adopted from DeepLab-v2~\cite{chen2017deeplab} with a ResNet-101 backbone.

To highlight the effectiveness of the proposed approach, we make a comparison of experiments from two different annotation strategy: random selection and active selection. The random selection strategy is used to select 10 patches, within each patch 5 points are selected for human annotation. Different from it, active selection strategy is used to select the 10 patches with highest mean entropy value, within each patch 5 points are selected for human annotation. Overall, our model achieves $51.5\%$ in mean IoU by using active selection. In the mean time, our model's performance only reach to $48.7\%$ in mean IoU by using random selection. This experiment demonstrates that the proposed active selection strategy outperforms the random selection strategy in the label acquisition stage.
To present the relevance of the proposed approach, we also conduct a comparison with self-training and full labeling method, as shown in Table~\ref{tab:gta5}. In the experiment of self-training method, we transfer the pseudo labels generated from UDA to the second stage of domain adaptation, without using any human annotations. In the experiment of full labeling method, we still generate 5 random points from each patch and we provide for all points human annotations, without using pseudo labels from UDA. By only using self-training method, the model only achieve $47.7\%$ in mean IoU, while the baseline model IntraDA reach its performance to $46.5\%$. It shows that self-training method is a less effective method to minimize the domain shift in this problem setting. The full labeling achieves up to $60.1\%$, functioning as a upper bound.

\begin{figure*}[ht!]
    \centering
    \includegraphics[width=0.94\textwidth]{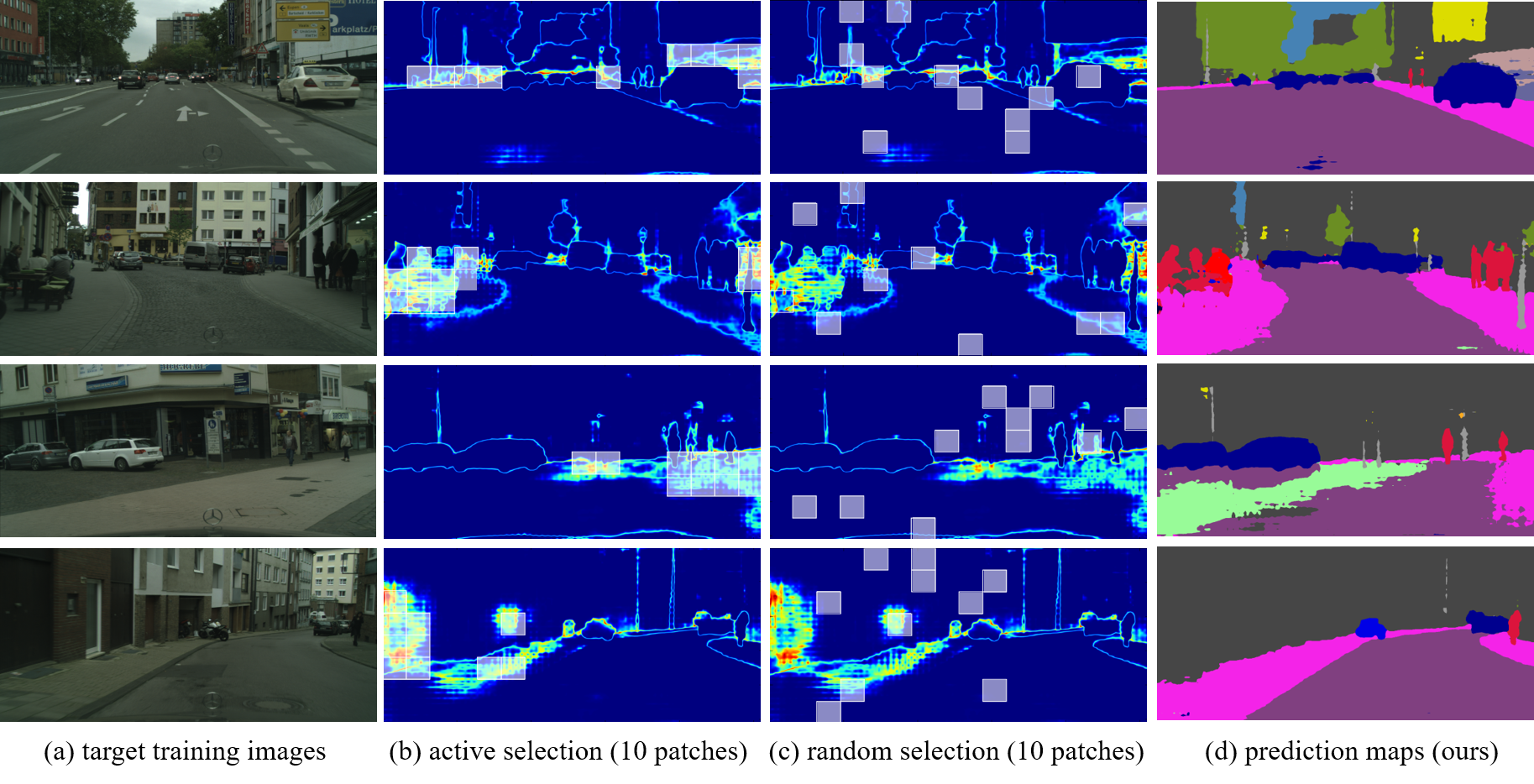}
    \caption{The examples of selection using active and random strategy in the evaluation for GTA5 $\to$ Cityscapes. (a) are the images from Cityscapes training set. (b) are the selected 10 patches from the entropy maps (from UDA) based on active strategy, shown as white rectangles. (c) are used to present the 10 selected patches from the entropy maps (from UDA) based on random strategy, also shown as white rectangles. (d) are the prediction maps generated from the proposed model using active selection. Note that in each selected patch, 5 random points are randomly generated for oracle annotation.}
    \label{fig:qualitative_2}
\end{figure*}

\vspace{7mm}

\noindent \textbf{Analysis of Hyperparameter $K$} We conduct an ablation study on the hyperparameter $K$ in our experiment of GTA5 $\to$  Cityscapes. In Table~\ref{tab:K}, different values of $K$ are used in the label acquisition system for patch selection. Remind that in the implementation process, the entropy map is divided by a grid of $16\times 8$. Thus the maximum value of $K$ is 128. For all experiments, we select patches with top $K$ highest mean entropy, and from each patch we randomly generate 5 points for oracle to annotate. It is obvious to discover that with the value of $K$ increase, the proposed model's performance also increase. This is because more supervision from human annotations are used during the training process.\\

\vspace{-3mm}

\noindent \textbf{SYNTHIA.} So as to show the effectiveness of the proposed approach, we also conduct the experiments related to the adaptation from SYNTHIA $\to$ Cityscapes. In these experiments, we also adopt the same segmentation model based on DeepLab-v2~\cite{chen2017deeplab}. The proposed approach is evaluated on both 16-class and 13-class baselines. According to Table~\ref{tab:synthia}, the proposed approach with active selection strategy of 10 patches achieves $55.8\%$ and $61.3\%$ in mean IoU on 16-class and 13-class baselines, respectively. However, our approach with random selection strategy of 10 patches achieves $49.8\%$ and $55.4\%$ in mean IoU on both two baselines. These experimental results suggest that our approach with active selection achieves higher performance than random selection strategy. 

In addition, we also presents experimental results of self-training and full labeling method in Table~\ref{tab:synthia}. In the self-training experiment, the proposed model achieves $42.0\%$ and $49.3\%$ of mean IoU in both 16-class and 13-class baselines. However, IntraDA~\cite{pan2020unsupervised} reaches its performance up to $41.7\%$ and $48.9\%$ in both baselines. It shows that the self-training method generates marginal improvements of performance. The full labeling method achieves $60.7\%$ and $66.2\%$ in both baselines, functioning as a upper bound.

\vspace{-2mm}

\section{Conclusion}
In this work, we propose a new domain adaptation framework for semantic segmentation with annotated points via active selection. First,  we conduct an unsupervised domain adaptation of the model;from this adaptation, we use an entropy-based uncertainty measurement  for  target  points  selection. Finally, to  minimize  the  domain  gap,  we  propose  a  domain  adaptation framework utilizing target points annotations. We present the experiments on synthetic data to real data in traffic scenarios. Experimental results on benchmark datasets shows the effectiveness of our approach against other domain adaptation approaches. 

{\small
\bibliographystyle{ieee_fullname}
\bibliography{egbib}
}

\end{document}